\documentclass[conference]{IEEEtran}
\IEEEoverridecommandlockouts
\usepackage{cite}
\usepackage{color}
\usepackage{setspace}
\usepackage{amsmath,amssymb,amsfonts}
\usepackage{textcomp}
\usepackage{float}
\usepackage{xcolor}
\usepackage{subfigure}
\usepackage{graphicx}
\usepackage{algorithm}
\usepackage{algorithmicx}
\usepackage{algpseudocode}

\def\BibTeX{{\rm B\kern-.05em{\sc i\kern-.025em b}\kern-.08em
    T\kern-.1667em\lower.7ex\hbox{E}\kern-.125emX}}
\begin{document}

\title{An Adaptive Device-Edge Co-Inference Framework Based on Soft Actor-Critic
\thanks{This work was supported in part by the National Key R\&D Program of China (No. 2021YFB3300100) and the National Natural Science Foundation of China (No. 62171062).}
}

\author{
\IEEEauthorblockN{Tao Niu\IEEEauthorrefmark{1}, Yinglei Teng\IEEEauthorrefmark{1} {\em IEEE Senior Member}, Zhu Han\IEEEauthorrefmark{2} {\em IEEE Fellow}, and Panpan Zou\IEEEauthorrefmark{1}}
\IEEEauthorblockA{\IEEEauthorrefmark{1}\textit{Beijing University of Posts and Telecommunications, Beijing, China} \\
\IEEEauthorblockA{\IEEEauthorrefmark{2}\textit{University of Houston, Texas, USA}}
Email: tasakim@bupt.edu.cn, lilytengtt@bupt.edu.cn}
}

\maketitle

\begin{abstract}
Recently, the applications of deep neural network (DNN) have been very prominent in many fields such as computer vision (CV) and natural language processing (NLP) due to its superior feature extraction performance. However, the high-dimension parameter model and large-scale mathematical calculation restrict the execution efficiency, especially for Internet of Things (IoT) devices. Different from the previous cloud/edge-only pattern that brings huge pressure for uplink communication and device-only fashion that undertakes unaffordable calculation strength, we highlight the collaborative computation between the device and edge for DNN models, which can achieve a good balance between the communication load and execution accuracy. Specifically, a systematic on-demand co-inference framework is proposed to exploit the multi-branch structure, in which the pre-trained Alexnet is right-sized through \emph{early-exit} and partitioned at an intermediate DNN layer. The integer quantization is enforced to further compress transmission bits. As a result, we establish a new Deep Reinforcement Learning (DRL) optimizer-Soft Actor Critic for discrete (SAC-d), which generates the \emph{exit point}, \emph{partition point}, and \emph{compressing bits} by soft policy iterations. Based on the latency and accuracy aware reward design, such an optimizer can well adapt to the complex environment like dynamic wireless channel and arbitrary CPU processing, and is capable of supporting the 5G URLLC. Real-world experiment on Raspberry Pi 4 and PC shows the outperformance of the proposed solution.
\end{abstract}


\section{Introduction}
In recent years, a series of AI based applications have been integrated into a variety of scenarios, such as image recognition \cite{b1}, natural language processing \cite{b2}, and intelligent voice assistant \cite{b3}. Typical network structures such as Convolutional Neural Network (CNN) and Deep Neural Network (DNN) can extract high-dimensional features of data, and thus have powerful processing performance. However, the breakthrough performance is at the cost of high computational complexity, which dramatically increases the inference latency and energy consumption of forward propagation. Hence, an unavoidable problem is how to efficiently deploy these deep models, especially in Internet of Things (IoT) scenarios. 

In order to solve the contradiction between huge resource demand of DNNs and resource shortage of devices, one traditional method is to unload data to a powerful server and return final results after computing. However, the rapid growth of intelligent devices will lead to huge pressure for cloud/edge computing. As the data size increases, communication becomes the bottleneck that limits the real-time data uploading. For example, the size of an image is usually about 2-10 MB, which takes one second or more to upload to an edge server with limited bandwidth, and thus is difficult to meet the requirements of 5G Ultra-reliable and Low Latency Communications (URLLC). Besides, such a centralized computation pattern requires data collection on a single point, which brings user data security risks once the cloud becomes unreliable \cite{b4}. Another solution is to run a lightweight version of the original deep network models on devices, usually compressed by {\em network pruning} or {\em quantization} \cite{b5}. However, the accuracy loss of these lightweight models is fixed when it is compressed that users can hardly shift the compromise between accuracy and execution latency for various applications.

To overcome the above challenges and deploy DNNs efficiently, we propose an device-edge collaborative optimization framework, which combines model partition with multi-branch selection and accelerates DNNs inference through an optimization design. By joint consideration of latency and energy consumption, we can deploy original models between the mobile device and the edge server while ensuring accuracy as much as possible. Moreover, our method can adaptively responds to dynamic environmental factors such as channel fluctuation, network bandwidth and CPU processing speed, etc. Specifically, the main contributions are summarized as follows:

\begin{itemize}
\item We study and implement different acceleration methods of DNNs under the device-edge framework. Based on measurement of \emph{early-exit}, \emph{model partition} and \emph{data quantization}, we reveal that there is a significant room for inference acceleration by combining these methods.
\end{itemize}

\begin{itemize}
\item Based on our insights above, we design a dynamic device-edge collaborative acceleration framework by adopting a novel deep reinforcement learning (DRL) method-Soft Actor-Critic for discrete (SAC-d) to make implementation decision on the exit point, partition point and quantization bits. Unlike previous studies \cite{b6}\cite{b7} that only consider the latency of DNNs inference, we treat the accuracy and latency adaptively in the optimization objectives through the hyperparameter setting.
\end{itemize}

\begin{figure*}[htbp]
\centerline{\includegraphics[width=6.35in]{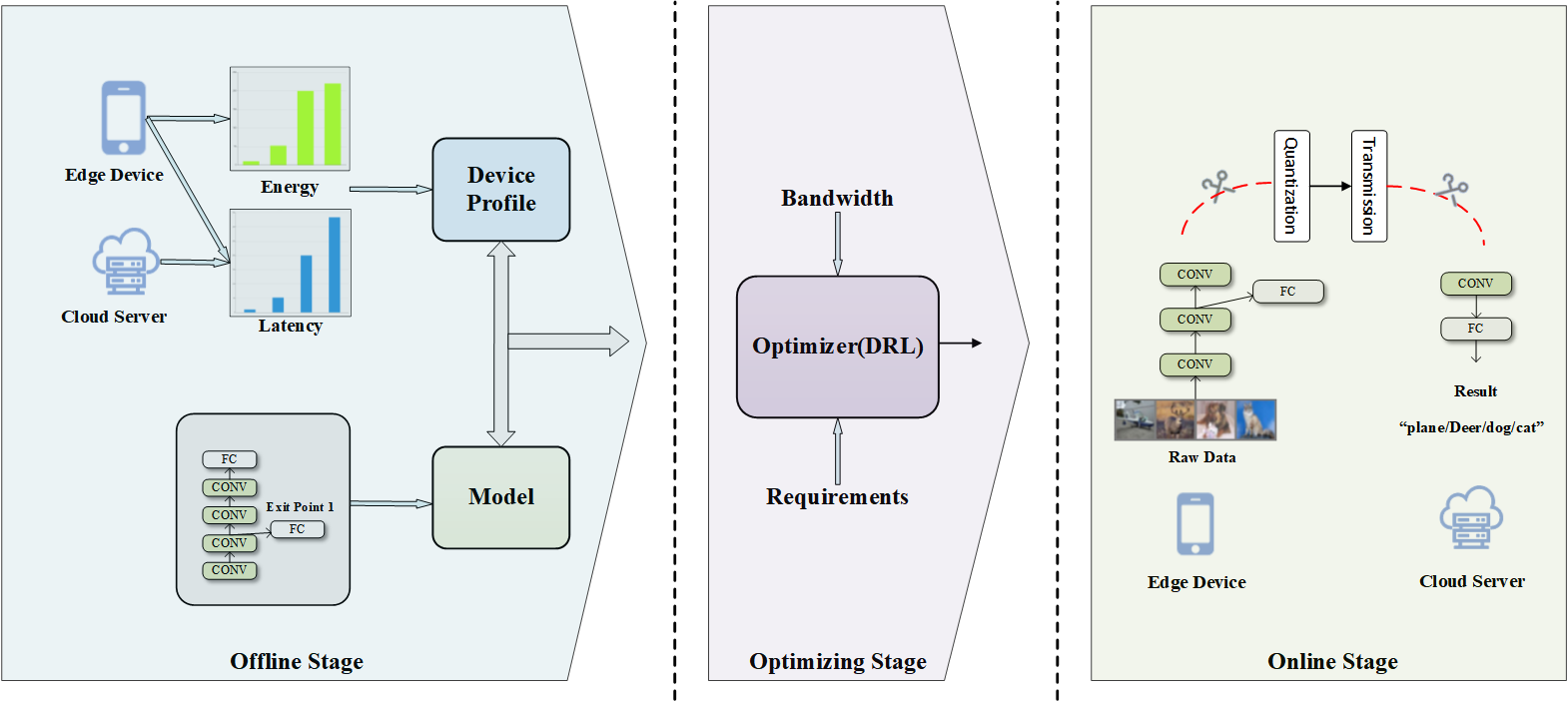}}
\caption{Overview of the proposed framework. Our framework consists of three stages (i) Offline Stage: Train the multi-branch network and measure data for optimization. (ii) Optimizing Stage: Use DRL to give the strategy about the exit point, partition point and quantization bit once requirements and bandwidth are set. (iii) Online Stage: Deploy models between the edge server and device based on the given strategy.}
\label{fig1}
\end{figure*}

\begin{itemize}
\item We carry out real-world experiments on Cifar-10 dataset by using Deep Q Network (DQN) and SAC-d. Experiment results show that the proposed framework can accelerate DNNs inference up to 4$\times$ and guarantee the performance even under limited bandwidth.
\end{itemize}

The rest of the paper is organized as follows. We discuss
related work in Sec. \ref{sec:II}. We present our framework details and formulation in Sec. \ref{sec:III}. We evaluate our method in Sec. \ref{sec:IV}, and finally conclude the paper in Sec. \ref{sec:V}.

\section{Related Work}
\label{sec:II}
In order to deploy pre-trained DNNs on different devices, researchers have proposed different deployment solutions. In this section, we give a comprehensive description on these techniques.
\subsection{Model Partition}

Model partition refers to dividing DNNs into two parts, e.g. between the edge and the device, with the computation-intensive part executed on the powerful server and the rest running locally. The mixed use of computation resource optimizes latency and energy consumption, as well as releases calculation pressure on devices, and thus obtains better inference performance.

Based on the current situation of cloud-based methods, Neurosurgeon \cite{b8} proposes the idea of DNN partition, which is a dynamic architecture that can automatically divide DNNs between the edge and cloud. Considering from both latency and energy efficiency aspects, Neurosurgeon proposes a regression-based method to estimate the latency of each layer in the DNN model and return an optimal partition point. JALAD \cite{b9} solves the optimization of end-to-end latency as an Integer Linear Programming (ILP) problem through modeling and obtains a reasonable partition scheme for different DNNs under dynamic environmental conditions. For the Industrial Internet scenario, Boomerang \cite{b6} uses DQN to make decisions on the partition scheme based on latency requirements and bandwidth conditions under the device-edge framework. However, a large amount of intermediate data is required to transmit in Boomerang, which may fail to cope with complex models.

\subsection{Early-Exit}
A model with high accuracy usually has a deep structure, which does not suit for local inference due to large resource consumption. Early-exit \cite{b10} is a special multi-branch structure based on the idea that shallow layers of the network can correctly classify most of the dataset. Traditional chain structures are added with side branches, from which classification results are allowed to output with different confidence. Then, by using entropy as the confidence criterion, it can achieve on-demand latency optimization.

DeepIns \cite{b11} proposes a manufacture inspection system for smart industry using early-exit. In DeepIns, devices are responsible for data collection, the edge server acts as the first exit point and the cloud data center acts as the second exit point. Edgent \cite{b9} combines model partition and early-exit, using collected device data to train a regression model to predict runtime of each layer. During inference, each partition point is traversed along branches to find an optimization solution, so as to achieve the effectiveness of balancing latency and accuracy. However, Edgent is not an industry-oriented framework and does not consider expensive energy consumption in the situation of limited bandwidth.

\subsection{Data Quantization\&Encoding}
Due to the existence of the convolutional layer, the shallow part of the network usually increases the dimensionality of the feature map, resulting in a much larger size of the intermediate data while partitioning models. In the pre-trained ResNet \cite{b12}, data size of some shallow layers can be 20$\times$ of the input data, which imposes an extra communication load. Data quantization$\&$encoding is one of the mainstream technologies for data compression. By using a more compact than 32-bit floating-point format to present layer inputs, weights, or both, it can not only reduce the memory occupation and transmission pressure, but also enhance the ability of privacy protection. Deep Compression \cite{b5} and Minerva \cite{b13} combine weight pruning and data quantization to enable fast, low-power and highly-accurate DNN inference. Based on model partition, JALAD \cite{b11} introduces min-max normalization and Huffman coding, horizontally analyzes the relationship between quantization bits and accuracy loss. \cite{b14} uses lossless/lossy encoding methods to compress data as much as possible while maintaining low loss of accuracy. 

\section{SYSTEM FRAMEWORK}
\label{sec:III}
In this section, we demonstrate our synthesized device-edge co-inference framework in detail. Fig. \ref{fig1} presents the workflow of our framework, illustrated in three stages: offline stage, optimizing stage and online stage. With the DRL optimizer, our framework can provide strategies of exit point, partition point and quantization bits dynamically according to the local transmission and energy  environment.

\subsection{Offline Stage}
During the offline stage, some initialization are performed:
i) To train a mutli-branch network and deploy it on both edge server and devices. ii) To measure layer latency and energy consumption for later optimization.

\begin{figure}[t]
\centerline{
\includegraphics[width=3.3in]{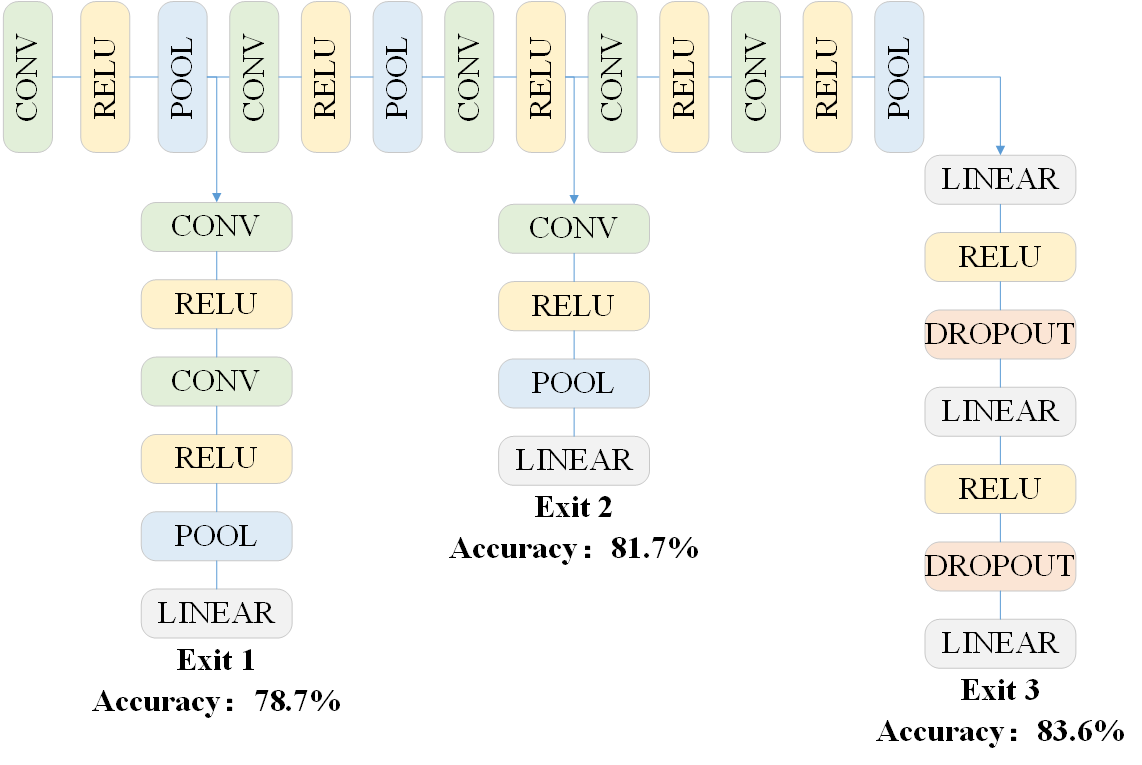}}
\caption{Trained multi-branch network based on Alexnet. Each exit possesses different accuracy.}
\label{fig2}
\end{figure}

Fig. \ref{fig2} shows a multi-branch structure based on Alexnet \cite{b15}, where Exit3 represents the main branch and the rest indicate side branches. As the exit point moves backward, each branch provides an increasing accuracy rate as the corresponding runtime and energy consumption increase as well. Once the multi-branch model is trained, it can be deployed between devices and edge servers for inference. Exit point can be selected according to the actual environmental factors and requirements.

\begin{figure}[t]
\centering
% \subfigure[Layer Latency]{
% \begin{minipage}{0.48\textwidth}
% \includegraphics[width=1\textwidth]{时延分布.png}
% \end{minipage}
% }
% \subfigure[Layer Energy Consumption]{
% \begin{minipage}{0.48\textwidth}
% \includegraphics[width=1\textwidth]{能耗分布.png}
% \end{minipage}}
% \caption{Layer Latency/Energy Consumption}
\includegraphics[width=3in]{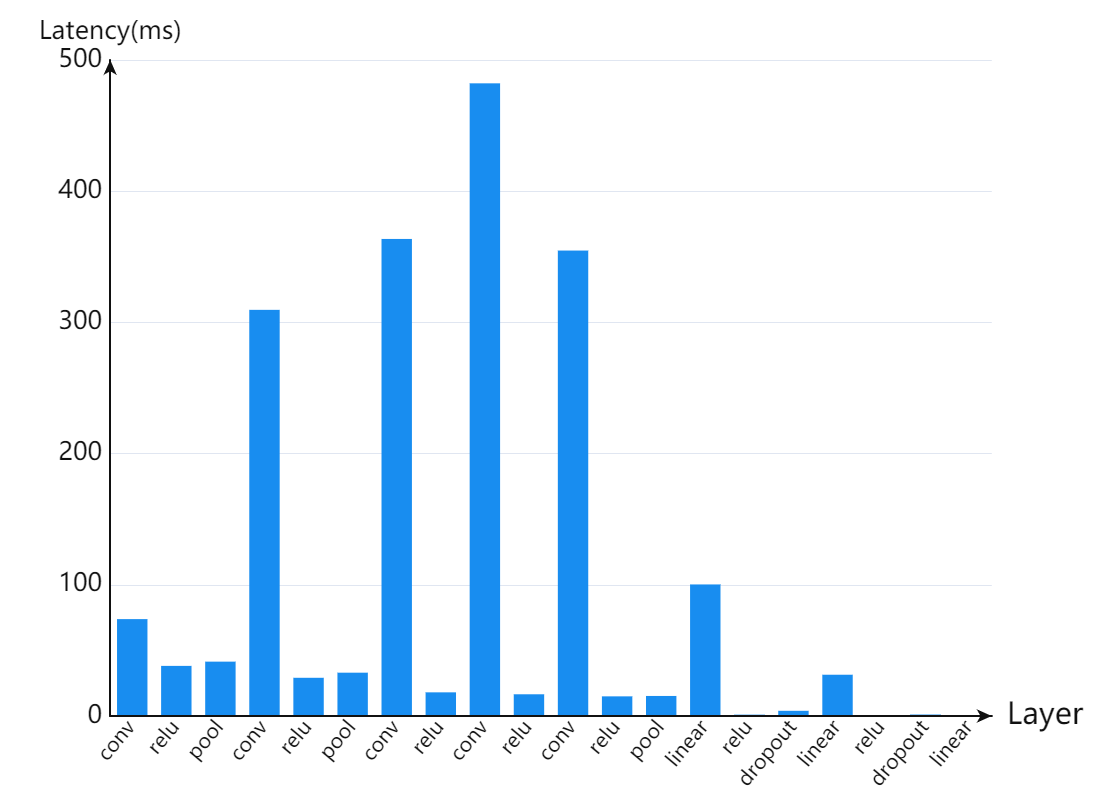}
\caption{Layer latency of multi-branch network when inferencing locally on Cifar-10 dataset (batchsize = 8).}
\label{fig3}
\end{figure}

In addition to the pre-trained multi-branch network, we need to measure the layer latency in advance. Previous work \cite{b7} uses a regression model to predict layer latency, but due to factors such as CPU resource scheduling, layer latency of the same device usually fluctuates. Therefore, it is difficult to obtain the accurate prediction by a regression model trained with measured data. By taking average of multiple measurements, we can obtain stable latency of each layer on devices and edge servers.

Former studies usually only consider layers with large FLOPs such as convolutional layers and fully connected layers while partitioning. However, as shown in Fig. \ref{fig3}, we find that although latency of other types of layers is relatively small, the accumulation of it is not neglectable. Hence, we measure the latency of each layer in order to optimize the overall latency as much as possible.

Other than accuracy and latency, we also take energy consumption of devices as optimization constraint. Computation energy consumption on devices is usually composed of multiple tasks, meanwhile, the transmission energy consumption is unstable due to the instantaneous fluctuations of the network. Such two factors together make it difficult to measure the total energy cost for a particular deep learning task. For tractability, we follow the theoretical modeling in \cite{b17} and calculate energy consumption as follows,
\begin{equation}
{E^{{compute}}} = \sum\limits_{{\rm{i}} = 1}^n {{k_0}{f^2}{O_i}{X_p}}\label{eq1},
\end{equation}
where $n$ is the number of network layers, ${k_0}$ is a constant related to the device, $f$ is the CPU frequency, ${O_i}$ is the computational intensity of each layer, and ${X_p}$ is the data size to be processed.

The transmission energy consumption can be written as,
\begin{equation}
{E^{transmission}}{\rm{ = }}\frac{{P{X_t}}}{{B{{\log }_2}\left( {1 + P*\gamma } \right)}}\label{eq2},
\end{equation}
where ${\gamma}$ indicates the signal to interference plus noise ratio, $P$ represents the transmission power between the device and edge server, $B$ is the available bandwidth, and ${X_t}$ is the data size to be transmitted.

Denote ${e}$ as the total energy consumption, the total energy consumption of inference can be written as,
\begin{equation}
{e = {E^{transmission}} + {E^{compute}}}\label{eq3}.
\end{equation}

Furthermore, to alleviate high communication overhead caused by excessively large intermediate data, we use the following formula to quantify it, which has been proven feasible for compressing the parameters \cite{b14}. By mapping 32-bit floating-point number ${x_i}$ to integer space of $[0, c]$, we can greatly reduces its size,
\begin{equation}
\overline {x_i}  = round\left(\frac{{(c - 1)({x_i} - \min (\bf{x}))}}{{\max (\bf{x}) - \min(\bf{x})}}\right)
\label{eq4}.
\end{equation}

\begin{figure}[t]
\centering
\includegraphics[width=3.3in]{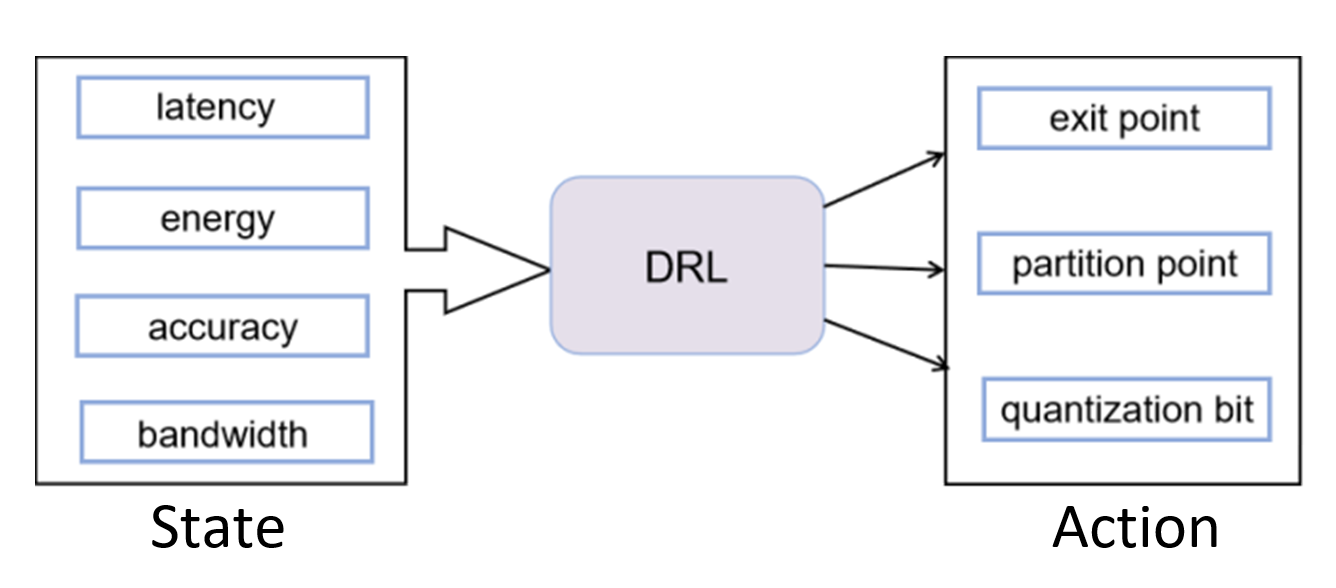}
\caption{The workflow of DRL optimizer.}
\label{fig4}
\end{figure}

After quantization, data becomes rather sparse, which contains a large number of zero values while most of the rest tend to zero. Therefore, intermediate data can be further compressed by encoding. We use the commonly used Huffman coding in our experiment, and results show that our data compression method can compress data to 1/10 or less. Nevertheless, quantization unavoidably causes a certain loss to the overall accuracy. Therefore, the accuracy drop should be included in our optimization when combined with model partition. We measure the accuracy loss of compressing feature maps of different partition points, which will be analyzed later.

\subsection{Optimization Stage}
Different from \cite{b6}, we do not take an exhaustive method to traverse all the possibilities after observing the bandwidth between the device and the edge server, since it is not efficient for complex network structures. By introducing DRL as an optimizer, we can make end-to-end decision with respect to exit point, partition point and quantization bit, as shown in Fig. \ref{fig4}. 

Denote $T$, $e$, $acc$ and $B$ as the latency, energy consumption, accuracy and bandwidth, respectively. Then at step $t$, we can express the environment state as,
\begin{equation}
{\mathbf{s}_t} = \{ {T_t},{e_t},{acc_t},{B_t}\} \label{eq5}.
\end{equation}

Denote $ep \in \{ 1,2,3\}$, $pp \in \{ 0,...,20\}$ and $c \in \{ 8,12,16\}$ as the exit point, partition point and quantization bits, respectively. Then the action selected by the agent at step $t$ is given by,
\begin{equation}
{\mathbf{a}_t} = \{ e{p_t},p{p_t},{c_t}\} \label{eq6}.
\end{equation}

\begin{itemize}
\item Soft Actor-Critic (SAC) for discrete action
\end{itemize}

Soft Actor-Critic for discrete action (SAC-d) \cite{b16} is an off-policy actor-critic method with soft policy updating based on the maximum entropy RL framework. Unlike the actor-critic architecture, the SAC updates policy $\pi$ through maximizing information entropy $H$ of state apart from the conventional cumulative rewards $r$, as described below,
% The overall algorithm is shown in Algorithm \ref{Algorithm 1}.
\begin{equation}
{\pi ^{\rm{*}}}{\rm{ = }}\mathop {\arg \max }\limits_\pi  \sum\limits_{t = 0}^T {{\mathbb{E}_{({\mathbf{s}_t},{\mathbf{a}_t})\sim{\tau _\pi }}}} \left[{\gamma ^t}(r({\mathbf{s}_t},{\mathbf{a}_t}) + \alpha H(\pi (.|{s_t})) \right] \label{eq7},
\end{equation}
where $\gamma$ is the discount factor and $\alpha$ is the entropy temperature that balance reward and entropy.

SAC uses one neural network with parameter $\theta$ for Q-value ${Q_\theta }$ and another neural network parameterized by $\phi$ for the policy ${\pi _\phi }$. To update these parameters, SAC alternates between a soft policy evaluation and a soft policy improvement. At the soft policy evaluation step, $\theta$ is updated by minimizing the following cost function,
\begin{equation}
% {J_Q}(\theta ) = {{\mathbb{E}_{({\mathbf{s}_t},{\mathbf{a}_t})\sim D}}\left[\frac{1}{2}{({Q_\theta }({\mathbf{s}_t},{\mathbf{a}_t}) - \mathop \hat{Q} ({\mathbf{s}_t},{\mathbf{a}_t}))^2}\right]}
{{J_Q}(\theta ) = {\mathbb{E}_{({{\mathbf{s}}_t},{{\mathbf{a}}_t})\sim{D}}}\left[ {\frac{1}{2}{{({Q_\theta }({{\mathbf{s}}_t},{{\mathbf{a}}_t}) - \hat Q({{\mathbf{s}}_t},{{\mathbf{a}}_t}))}^2}} \right]}, 
\label{eq8}
\end{equation}
where
\begin{equation}
\label{eq9}
\begin{array}{l}
\hat Q({{\mathbf{s}}_t},{{\mathbf{a}}_t}) = r({{\mathbf{s}}_t},{{\mathbf{a}}_t}) + \\ \gamma {\mathbb{E}_{({{\mathbf{s}}_{t + 1}},{{\mathbf{a}}_{t + 1}})\sim D'}}\left[ {\hat Q({{\mathbf{s}}_{t + 1}},{{\mathbf{a}}_{t + 1}})} \right. - {\alpha \log ({\pi _\phi }({{\mathbf{a}}_{t + 1}}|{{\mathbf{s}}_{t + 1}}))}\Big].
\end{array}
\end{equation}

$D$ is an experience replay buffer, $D'$ is composed of the state transition distribution and the agent, and ${Q_\theta }$ denotes the target $Q$ function whose parameters are periodically updated from the learned $\theta$. Since the action space is discrete, \cite{b16} fully recovers the action distribution instead of forming a Monte-Carlo estimate, and calculates the expectation directly to reduce the variance of estimation at the soft policy  improvement step. The policy $\pi$ with its parameters $\phi$ is updated by minimizing the following objective,

\begin{equation}
{J_\pi }(\phi ) = {\mathbb{E}_{{\mathbf{s}_t}\sim D}} \left[{\pi _t}{({\mathbf{s}_t})^T}[\alpha \log ({\pi _\phi }({\mathbf{s}_t})) - {Q_\theta }({\mathbf{s}_t})] \right] \label{eq10}.
\end{equation}

Similarly, the temperature objective is changed as,
\begin{equation}\label{eq11}
J(\alpha ) = {\pi _t}{({\mathbf{s}_t})^T} \left[ \alpha \log ({\pi _t}({\mathbf{s}_t})) + \mathbf{\mathop H\limits^ -}  \right],
\end{equation}
where ${\mathbf{\mathop H\limits^ -}}$ is a constant vector represents the target entropy.

In order to put more emphasis on latency when bandwidth is limited and to pursue accuracy when it is sufficient, hyperparameters $a$ and $b$ are introduced in our reward function, then at step $t$, the reward ${r_t}$ can be defined as below,
\begin{equation}
% r = a *accuracy + b /latency \label{eq12}.
{r_t} = a *{acc_t} + b /{T_t} \label{eq12}.
\end{equation}

\begin{figure*}[htbp]
\centerline{
\subfigure[Comparison]{\includegraphics[width=0.48\textwidth]{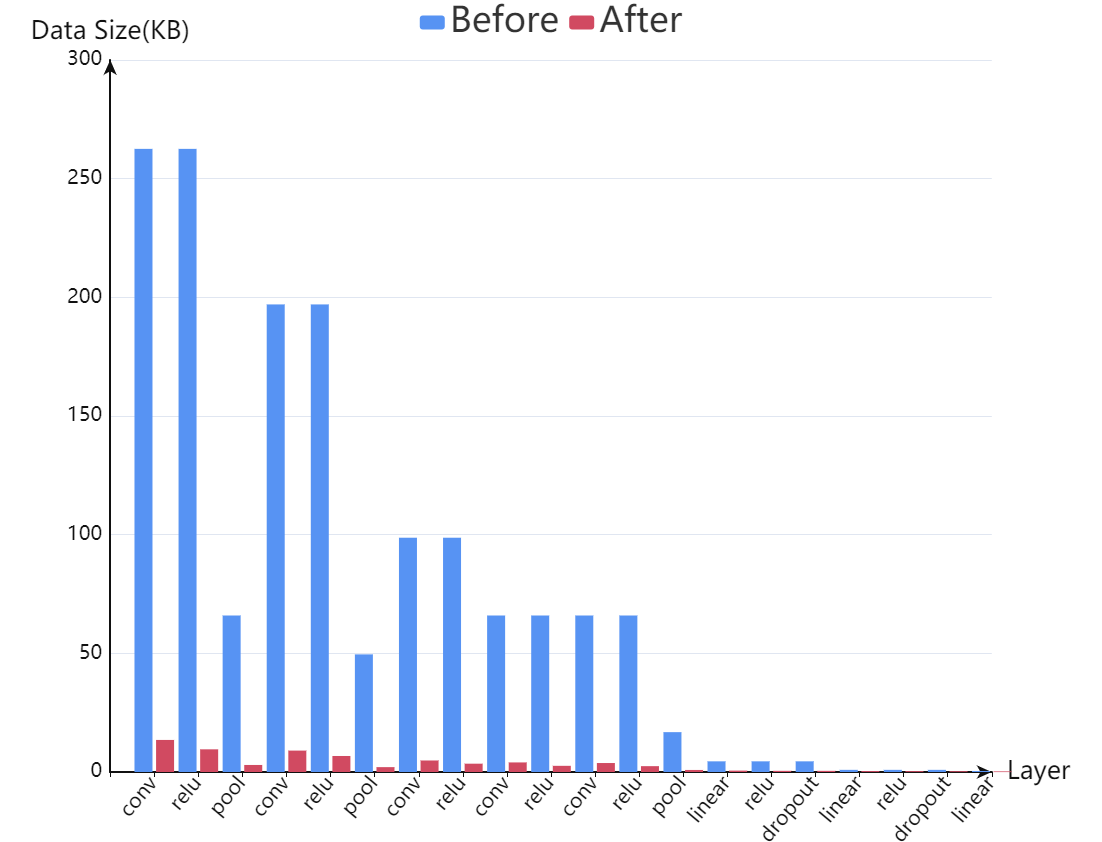}}
\subfigure[Accuracy Loss]{\includegraphics[width=0.48\textwidth]{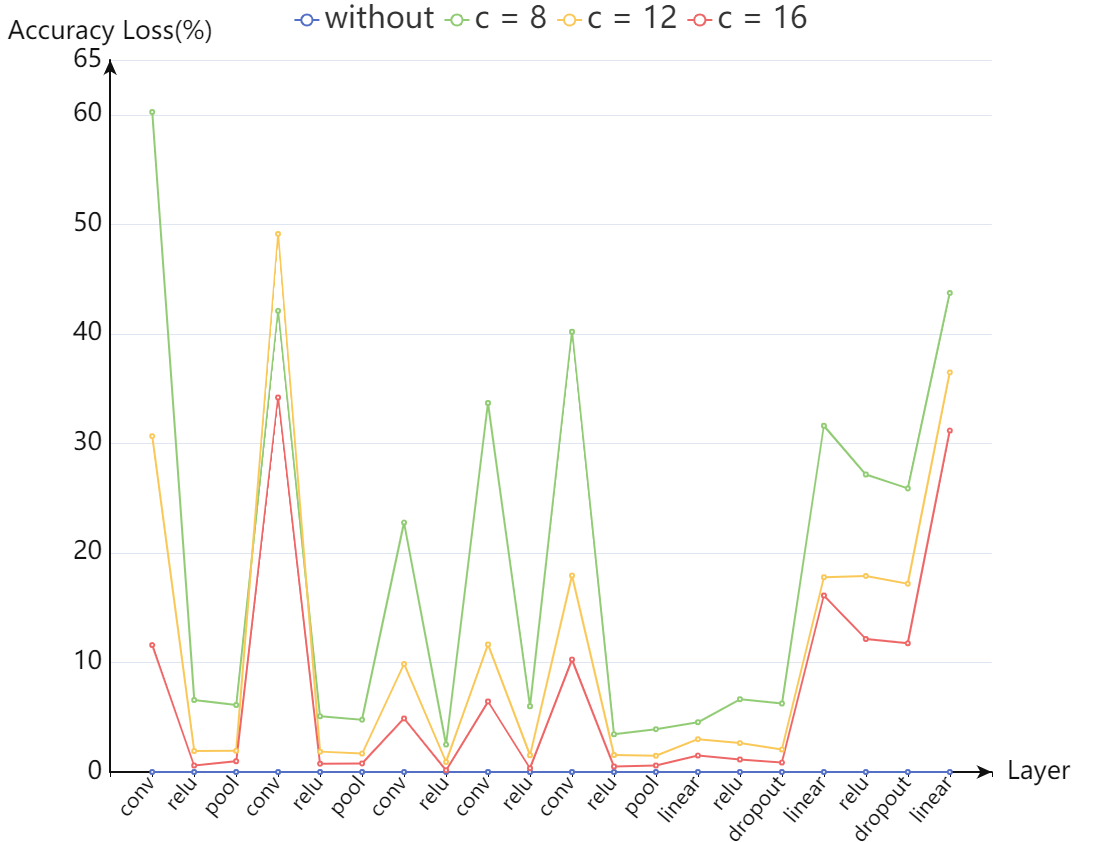}}
}
\caption{Compression performance of data quantization at different partition points.}
\label{fig5}
\end{figure*}

To provide an automatic fine-tune objective that can guide agent to react dynamically to the bandwidth, we first standardize accuracy and latency to the same magnitude. Then the hyperparameter $a$ is set to a logarithmic form which varies with the bandwidth, as shown below,
\begin{equation}
a  = n*\ln \left(1 + \frac{B}{s}\right) \label{eq13},
\end{equation}
where $n$ is the normalization coefficient that limits $a$ between $0$ and $1$, $s$ is used to describe the sensitivity of $a$ to different bandwidth, and $b$ is set to $1-a$ to balance the effect of accuracy and latency.

The entire DRL optimization process can be written as,
\begin{equation}
\begin{split}
&\max\quad \sum\limits_t {{r_t}}\\
% \sum\limits_t {{r_t}}\\
% r = a *accuracy + b /latency\\
&s.t.\quad {r_t} = \left\{ {\begin{array}{*{20}{c}}
{a*ac{c_t} + b/{T_t},}&{e \le \hat e,}\\
0&{else,}
\end{array}} \right.
% e\leq \hat{e},
\end{split}
\label{eq14}
\end{equation}
where $\hat{e}$ is the constraint of energy consumption of the end device.

\subsection{Online Stage}
During the online stage, based on the real-time uplink detection, the DRL optimizer can dynamically output an inference strategy and guide device-edge to execute the DNNs collaboratively. Our DRL optimizer can be executed either on the edge server or the device since its runtime is quite short, i.e., less than 5ms.

\section{EXPERIMENT}
\label{sec:IV}
In this section, we carry out real-world experiments to evaluate the performance of our design. For comparison, the DQN is chosen as baseline and explored to test the effectiveness for our proposed co-inference pattern.

\subsection{Experiment Setup}\vspace{1mm}
\subsubsection{Model Structure}
Due to the popularization of CNN in image classification, we focus on the disassembled CNN structure. Specifically, a multi-branch network based on Alexnet is adopted and trained from scratch for 100 epochs on Cifar-10 dataset. As mentioned above, our network has three branches, each with 9/12/20 layers. During training, we set $batch\_size$ as 128, $learning\_rate$ as 0.01 with exponential decay. Accuracy of each branch increases as exit point moves backward, which is $78.7\%$, $81.7\%$ and $83.6\%$. Besides, we set the quantization bits $c$ as 4, 8 and 12 in order to shrink the action space.
\begin{figure*}[htbp]
\centering
\subfigure[]{\includegraphics[width=0.32\textwidth]{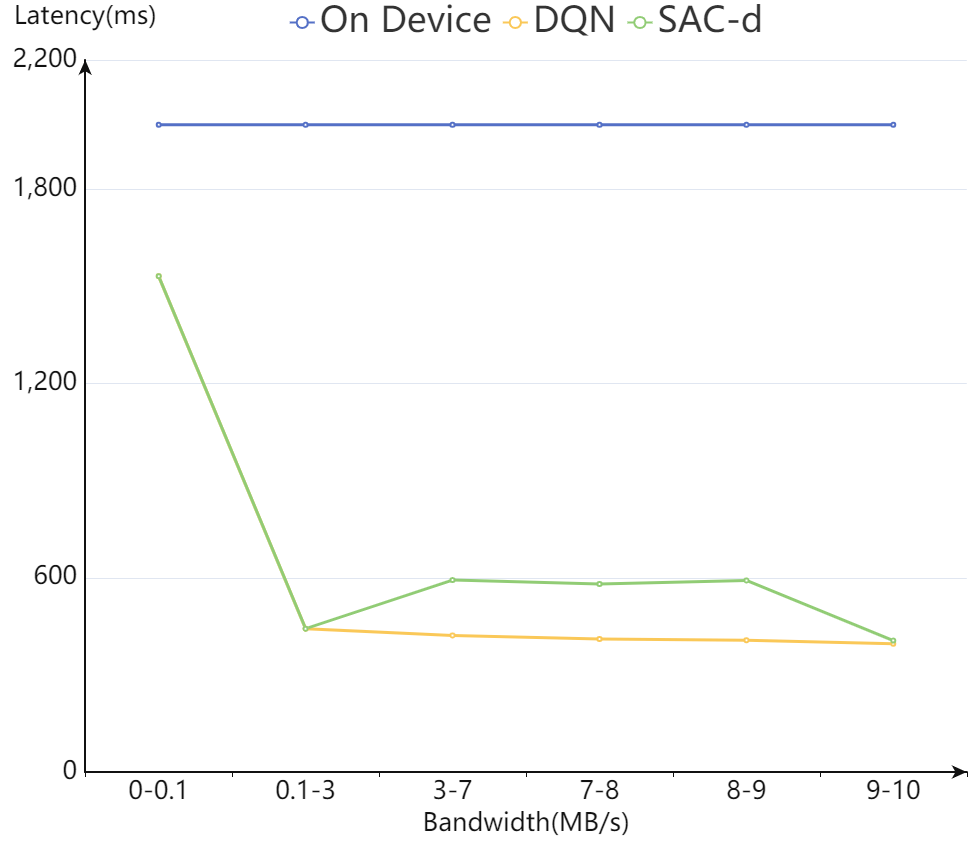}}
\subfigure[]{\includegraphics[width=0.32\textwidth]{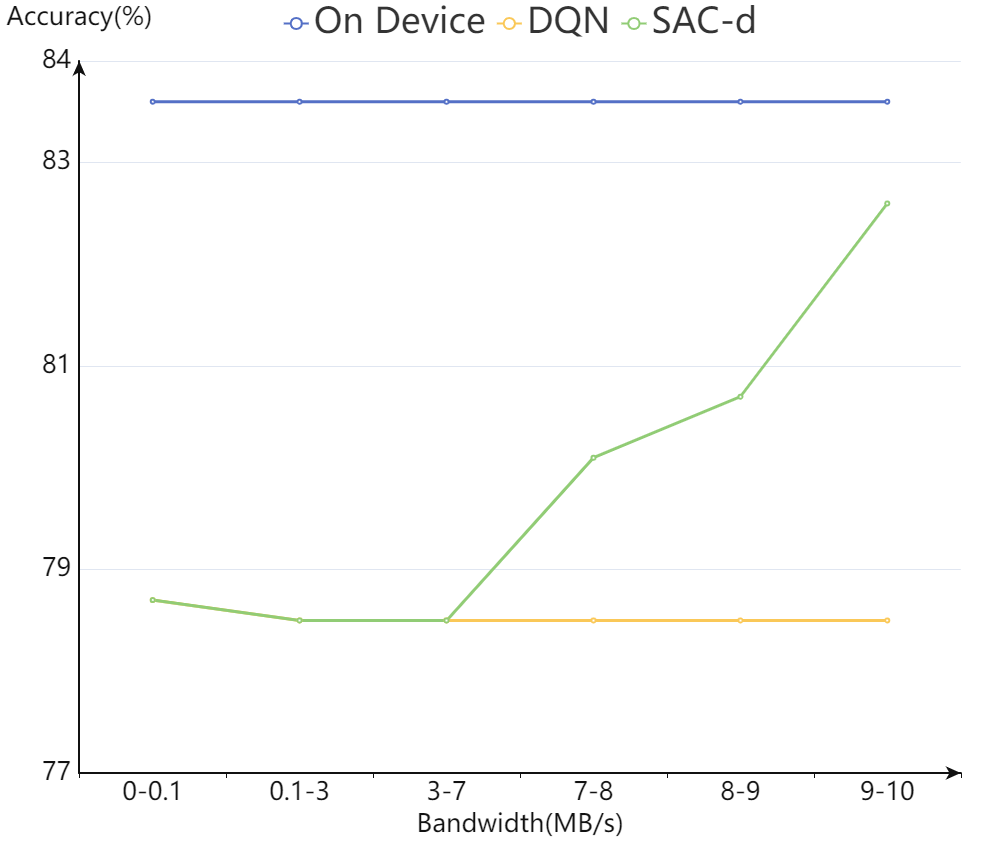}}
\subfigure[]{\includegraphics[width=0.34\textwidth]{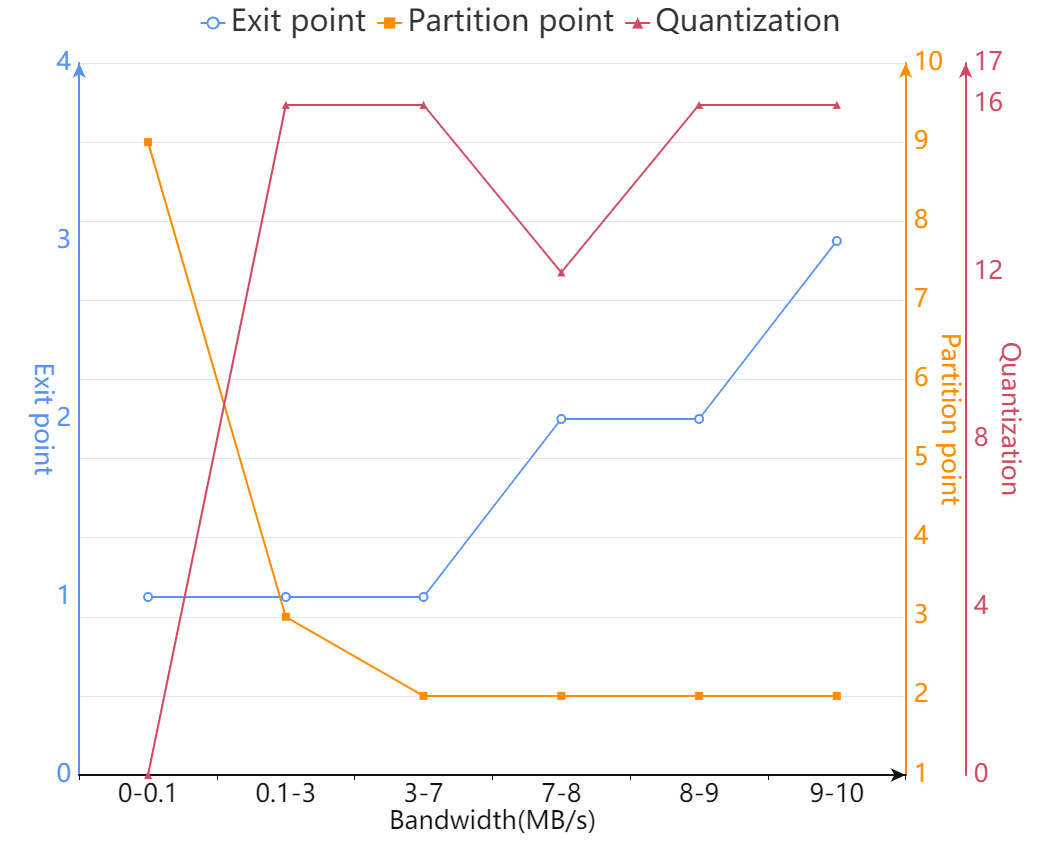}}
\caption{Evaluation results under different bandwidths. (a) Latency comparison. (b) Accuracy comparison. (c) Points selection result of SAC-d.}
\label{fig6}
\end{figure*}

\subsubsection{Real-world Experiment}
We choose Raspberry Pi 4 as the end device and a laptop as the edge server. Raspberry Pi 4 has a 4-core 64-bit cpu with a frequency of up to 1.5GHz and 8G of RAM, while the laptop is equipped with a 6-core, 12-thread i7-10710u and 16G of memory. Both devices use WiFi for connection, and the available bandwidth between the device and the edge server is controlled by the WonderShaper tool. During inference, to obtain a traceable CPU frequency, we set both laptop and Raspberry Pi to its maximum CPU frequency.

\subsubsection{Optimizer}
Both SAC-d and DQN consist of five fully connected layers. In terms of traning parameters, we set $\gamma$ = 0.99, $learning\_rate$ = 0.001. After collecting data such as latency, energy consumption in the offline stage, DRL optimizer is trained till convergence. Moreover, both multi-branch network and DRL in our framework are implemented by Pytorch.

\subsection{Experiment Result}
In our experiments, we deploy the pre-trained multi-branch network on Raspberry Pi and PC to verify whether our framework can dynamically make decisions that meet requirements under a given limited bandwidth. 
\subsubsection{Layer Latency/Energy Distribution}
Fig. \ref{fig3} shows the distribution of different layers on Raspberry Pi. It can be seen that the convolutional layers and fully connected layers take up most of the computing time. Although activation function layers and pooling layers take less time, the amount of them is non-negligible towards choosing the partition point, which makes it necessary to take them into consideration. 
\subsubsection{Impact of Data Quantization}
We adopt data quantization$\&$coding for intermediate data in order to alleviate limitation caused by bandwidth. Fig. \ref{fig5}(a) shows the difference after compression of different partition points, which clearly indicates the great reduction of data sizes. Even though quantization$\&$coding significantly compresses the size of data, the accuracy loss it causes cannot be ignored. Fig. \ref{fig5}(b) shows the impact of different quantization bits on accuracy. It can be seen that different quantization bits take the same trend, i.e., the higher quantization bits $c$ is, the lower the accuracy loss becomes. Meanwhile, for the output of convolutional layers and the fully connected layers, serious accuracy drop will be caused when quantifying. Contrarily, this is not obvious on the activation function layers and pooling layers, which indicates the partition point should generally be selected at these type of layers.

\subsubsection{Optimization Analysis}
Fig. \ref{fig6}(c) shows the strategy of SAC-d as we vary the bandwidth from 0 to 10 MB/s. When bandwidth is limited in 0-0.1 MB/s, communication overhead becomes the main bottleneck. Transmitting intermediate data causes huge latency since the first exit point is chosen, which means the entire DNN is processed on the device. As the bandwidth gradually increases, exit point remains unchanged while partition point moves forward and quantization bits is selected as the largest 16 bits, which indicates the computing power of device and communication overhead are both main bottlenecks in this situation. As the bandwidth further increases to 7 MB/s, exit point changes to 2, indicating that the bandwidth allows system to pursue higher accuracy. However, the communication overhead still accounts for a relatively large amount, and so quantization bits is selected as 12. When the bandwidth is between 8 and 9 MB/s, due to the further improvement of the bandwidth, the system can use 16 bits to compress data so as to pursue higher accuracy. When the network conditions are relatively good, e.g., the bandwidth is above 9 MB/s, the exit point changes to 3, which gives more allowance for inference accuracy.

Figs. \ref{fig6}(a) and (b) show the comparison of different optimizers on latency and accuracy. The curve "On Device" means that the original network is executed entirely on edge device without early-exit. As the curve of latency after optimization are all below that of on device, both DQN and SAC-d bring significant improvement with a little degradation on accuracy, which imply that they can well adjust to fluctuating network. It can be seen that although the latency performance of SAC-d is slightly worse than that of DQN in some region of bandwidth, DQN shows a trend of only focusing on latency rather than pursuing accuracy as well as latency in the same time, which indicates DQN cannot cope with too large action space. As the bandwidth no longer becomes system bottleneck, the accuracy of SAC-d is far more higher than that of DQN.
% In terms of latency, it can be seen that the performance of SAC-d is better than that of DQN, as DQN cannot cope with too large action space to pursue the lowest latency in different region of bandwidth. Also we can see from Fig. \ref{fig6}(b) that the accuracy of SAC-d is much better than that of DQN, which proves that our framework can optimize the accuracy and latency at the same time.

\section{Conclusion}
\label{sec:V}
This paper proposes a dynamic inference framework in an IoT environment, which integrates early-exit, model partition and data quantization$\&$coding. We study the trade-off between the model accuracy and execution latency when a deep neural network is decoupled to execute on an edge server and device. Based on the detected real-time bandwidth as inputs, the proposed DRL optimizer can output a strategy between the device and edge server that guide them execute DNNs. Experiments based on Raspberry Pi and PC prove the outperformance of this device-friendly co-inference framework on latency, energy consumption and accuracy, which makes it more suitable for mission-critical IoT cases.

% \section*{Acknowledgment}
% This work was supported in part by the National Key R&D Program of China (No. 2021YFB3300100) and the National Natural Science Foundation of China (No. 62171062).

\end{document}